\documentclass[journal]{IEEEtran}

\usepackage{bm}
\usepackage{graphicx}
\usepackage{xcolor}
\usepackage{caption}
\usepackage{booktabs}
\usepackage{url}
\usepackage{subfigure}
\usepackage{amsmath}
\DeclareMathOperator*{\argmax}{argmax}
\usepackage[noadjust]{cite}

\graphicspath{{figures/}}

\begin{document}

\title{Do Models Learn the Directionality of Relations?\\A New Evaluation: Relation Direction Recognition}

\author{Shengfei~Lyu, Xingyu~Wu, Jinlong~Li, Qiuju~Chen, and~Huanhuan~Chen,~\IEEEmembership{Senior~Member,~IEEE}

\thanks{© 20XX IEEE.  Personal use of this material is permitted.  Permission from IEEE must be obtained for all other uses, in any current or future media, including reprinting/republishing this material for advertising or promotional purposes, creating new collective works, for resale or redistribution to servers or lists, or reuse of any copyrighted component of this work in other works.}
}
\maketitle

\begin{abstract}
	Deep neural networks such as BERT have made great progress in relation classification.
	Although they can achieve good performance, 
	it is still a question of concern whether these models recognize the directionality of relations, 
	especially when they may lack interpretability.
	To explore the question, a novel evaluation task, called Relation Direction Recognition (RDR), 
	is proposed to explore whether models learn the directionality of relations. 
	Three metrics for RDR are introduced to measure the degree to which models recognize the directionality of relations.
	Several state-of-the-art models are evaluated on RDR. 
	Experimental results on a real-world dataset empirically indicate that there are clear gaps among them in recognizing the directionality of relations, 
	even though these models obtain similar performance in the traditional metric (i.e. Macro-F1).
	Finally, some suggestions are discussed to enhance models to recognize the directionality of relations 
	from the perspective of model design or training.
\end{abstract}

\begin{IEEEkeywords}
Relation Direction Recognition, Relation Classification, Relation Extraction.
\end{IEEEkeywords}


\section{Introduction}

\IEEEPARstart{R}{elation} classification, a supervised version of relation extraction, aims to predict a relation between given entities  in a sentence  
and is a crucial step to construct knowledge bases, 
benefiting many applications such as 
question answering \cite{zhao-2020-conditiion},
knowledge graph enrichment \cite{trisedya-etal-2019-neural}, 
and natural language generation \cite{cai-etal-2020-learning}.
For example, relation classification requires methods to classify the entities, \textit{Jack} annotated by \textit{e1} and 
\textit{Jackson} annotated by \textit{e2}, of the first 
sentence in Table \ref{tab:example} to the  relation \textit{Father-Child(e1,e2)}.

With the development of deep learning, 
many methods, based on CNN \cite{wang-etal-2016-relation}, 
RNN \cite{zhang-etal-2017-position}, 
GCN \cite{zhang-etal-2018-graph, guo-etal-2019-attention}, 
and pretrained language models \cite{baldini-soares-etal-2019-matching, wu-he-2019-enriching},  
can achieve good performance on relation classification. 
The performance is usually evaluated on the traditional metrics, such as accuracy, Macro-F1, and Micro-F1.

These neural models are regarded as black-boxes \cite{mehta-etal-2020-simplify}, 
meaning that they may be not interpretable.
Even though these models achieve the same performance, 
it is hard to claim that they are similar in understanding sentences or recognizing relations.
More explorations and evaluations are still required to better distinguish these models and understand their internal mechanisms.

Developing interpretable deep neural networks, which converts these black-boxes into white- or grey-boxes, is very valuable but difficult work.
It might take a number of researchers decades to address this open question.
Alternatively, an easy and feasible way to deepen the understanding of these black-boxes is to evaluate them from various aspects, 
which is studied in the paper.

\begin{figure}[tp]
	\centering
	\includegraphics[width=0.65\columnwidth]{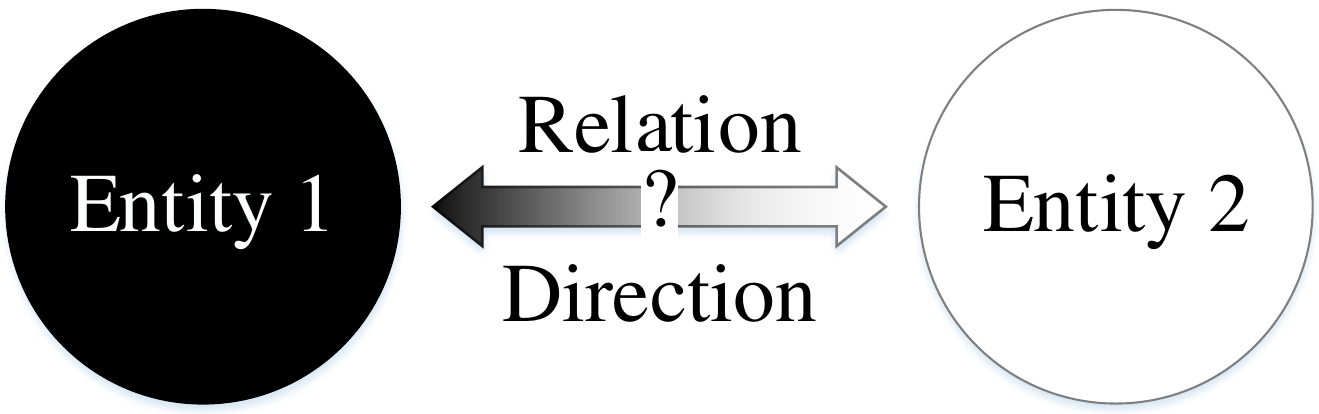}
	\caption{The directionality of a relation.}
	\label{fig:relation-direction}
\end{figure}

\begin{table*}[t]
	\centering
	\caption{Two samples that contain similar sentences and different relations due to the opposite entity annotation orders. }
	\begin{tabular}{cccc}
		\toprule
		Sample & Sentence & Ground-truth Relation & \\
		\midrule
		No. 1  & \textcolor{red}{\textless e1\textgreater \,  Jack \textless /e1\textgreater} \, is the father of \, \textcolor{blue}{\textless e2\textgreater \, Jackson \textless/e2\textgreater} &  Father-Child(\textcolor{red}{e1},\textcolor{blue}{e2}) \\ 
		\midrule
		No. 2  & \textcolor{blue}{\textless e2\textgreater \, Jack \textless /e2\textgreater} \, is the father of \, \textcolor{red}{\textless e1\textgreater \, Jackson \textless/e1\textgreater} &  Father-Child(\textcolor{blue}{e2},\textcolor{red}{e1}) \\ 
		\bottomrule
	\end{tabular}
	\label{tab:example}
\end{table*}

Some interesting phenomena can be explored in relation classification to access these uninterpreted models.
An obvious common sense about relations is that they are directional (Fig. \ref{fig:relation-direction}).
It is important for a model to recognize the directionality of a relation in addition to understanding its meaning.
The directionality of relations has a great influence on knowledge inference \cite{cai-eatl-2019-transgcn}.
As to the directionality of a relation, 
a model needs to capture the relative order of entities in a sentence.
It can deepen the understanding of a model by judging whether the model recognizes the directionality of relations.

The current evaluation for a model in relation classification aims to comprehensively assess its performance in distinguishing different relations.
Traditional metrics (e.g. Macro-F1) usually quantify the performance as a whole.
However, the current evaluation may be unable to directly judge whether models recognize the directionality of relations,
and the degree of recognition may not be explicitly reflected by the traditional metrics, 
since the evaluation and the metrics are not designed for this purpose. 
Therefore, it is a question of concern to discern and measure the degree to which models recognize the directionality of relations, 
especially when these models lack interpretability.

To explore this, a novel evaluation task, called Relation Direction Recognition (RDR), 
is proposed to evaluate if a model recognizes the directionality of relations in this paper. 
Under the RDR evaluation, a model that is trained once will be evaluated twice in a pair of paired test sets, 
which contain similar yet different samples that are specially designed for RDR.
For example, one of the paired sets embraces the first sample in Table \ref{tab:example} while the other set holds the second sample in Table \ref{tab:example}.
The results of a model on a pair of paired test sets are compared with each other to formulate the RDR evaluation.
Based on the difference or consistency of the two results, 
three metrics are introduced to measure the degree 
to which the model recognizes the directionality of relations. 

Several state-of-the-art methods in relation classification are evaluated under RDR 
on a real-world dataset.
Although these methods achieve similar performance in the traditional evaluation (i.e. Macro-F1),
they have clear gaps in the proposed metrics for RDR. 
In other words, these experimental results empirically indicate the differences among these  methods 
in recognizing the directionality of relations. 
By the RDR evaluation, 
these methods could be further understood especially when they are  uninterpretable. 

Additionally, some suggestions that aim to recognize the directionality of relations are discussed in the paper.
We observe that exploiting paired training sets can not only force a model to recognize the directionality of relations 
but also further improve the performance of the model.
Based on this observation, new state-of-the-art 
performance\footnote{  \url{https://paperswithcode.com/sota/relation-extraction-on-semeval-2010-task-8}{ Only the results on published papers are considered.}} 
is achieved on SemEval-2010 Task 8 \cite{hendrickx-etal-2010-semeval}.  
Besides, how much weight a model allocates to various input elements (i.e. tokens) is illustrated
to analyze the reason why the model can or cannot recognize the directionality of relations.

To recap, the main contributions of the paper are summarized as follows:
\begin{itemize}
	\item A new evaluation, called  relation direction recognition (RDR), is proposed to  evaluate whether models recognize the directionality of relations. 
	Furthermore, three metrics are introduced to measure the degree of recognition from various aspects.
	\item Experimental results empirically indicate that several state-of-the-art models for relation classification perform very differently 
	in recognizing the directionality of relations.
	\item Some suggestions are discussed to assist a model to recognize the directionality of relations.
	Based on these suggestions,  new state-of-the-art performance is achieved on SemEval-2010 Task 8.
\end{itemize}

The rest of this paper is organized as follows. The related
work is introduced in Section II. Section III
presents the proposed relation direction recognition and its three evaluation metrics. 
The experimental results
and discussion are presented in Sections IV and V, respectively.
Finally, Section VI concludes this paper.

\section{Related Work}

With the development of deep learning, 
various neural networks are exploited in relation classification. 

In the last decade, an active area of research was studied in relation classification by extracting features from sentences by
convolutional neural network (CNN) \cite{lecun-1998-cnn},
including non-attention \cite{zeng-etal-2014-relation,nguyen-grishman-2015-relation,xu-etal-2015-semantic,zeng-etal-2015-distant,dos-santos-etal-2015-classifying} and attention \cite{wang-etal-2016-relation} versions.
Especially, our work is inspired by the negative sampling strategy introduced in \cite{xu-etal-2015-semantic}, 
which treats samples with the opposite entity annotation order as  negative samples.
The main difference from \cite{xu-etal-2015-semantic} in the paper is that 
the positive samples are constructed and exploited to explore if models recognize the directionality of relations.

By using recurrent neural network (RNN) \cite{hochreiter1997long}, learning the hidden states of words in a sentence is also extensively studied   \cite{xu-etal-2015-classifying,xu-etal-2016-improved,miwa-bansal-2016-end,zhang-etal-2017-position,yu-etal-2019-beyond,lyu-etal-2020-auxiliary} in predicting relations. 
As a special type of RNN, echo state networks had been utilized for NLP tasks \cite{popov-etal-2019-etal-echo} and other applications \cite{chen-etal-2013-model, chen-etal-2014-learning}.

By exploiting graph convolutional network (GCN) \cite{kipf-2017-gcn}, the long-range dependency trees parsed from sentences are incorporated to learn relation representations for classification. 
A non-attention version \cite{zhang-etal-2018-graph} and an attention version \cite{guo-etal-2019-attention} are explored in relation classification. 
Besides, a novel paradigm, RElation Classification with ENtity Type restriction, RECENT\cite{lyu-chen-2021-relation}, is proposed and integrates GCN into  $\rm RECENT_{GCN}$, which outperforms significantly its base model GCN.

\begin{figure*}[t]
	\centering
	\includegraphics[width=2\columnwidth]{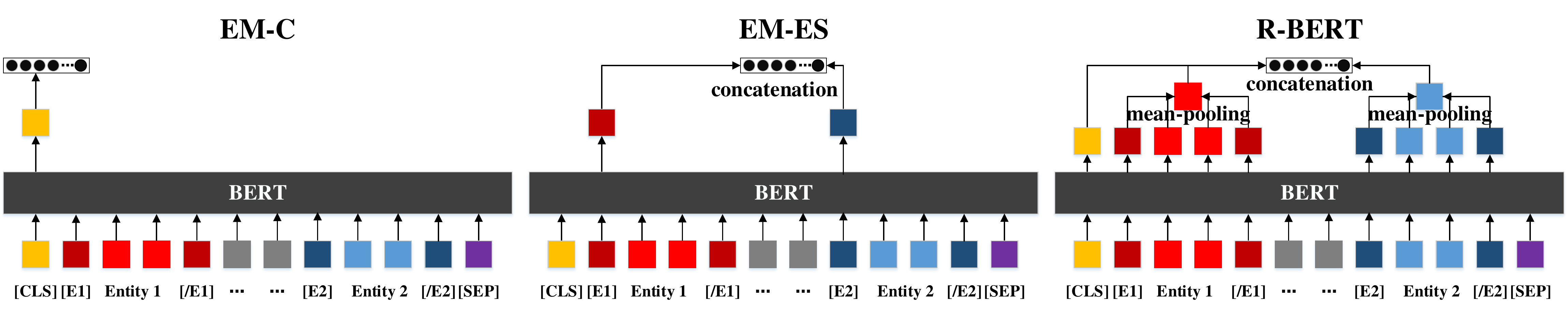}
	\caption{The architectures of the three methods for relation classification: \textit{EM-C}, \textit{EM-ES}, and \textit{R-BERT}.}
	\label{fig:r-bert}
\end{figure*}

More recently, pretrained language models (e.g. BERT \cite{devlin-etal-2019-bert}) have made a big breakthrough in a number of NLP tasks. 
BERT is pretrained on a very large corpus and can fine-tune in many downstream tasks.
For relation classification, several methods based on BERT are proposed to highlight entities in a sentence.
Baldini Soares et al. \cite{baldini-soares-etal-2019-matching} attempt six strategies to incorporate the information of entities.
One of them is \textit{entity markers and [CLS]} (EM-C) 
that introduces four tokens pinpointing entities' positions in sentences as input 
and uses the hidden state of the special token {[CLS]} of BERT as output to predict relations 
(the leftmost part in Fig. \ref{fig:r-bert}).
Another strategy, named \textit{entity markers and entity start} (EM-ES), 
adopts the same input as EM-C yet utilizes the concatenation of the hidden states of two \textit{entity start markers} as output to make predictions 
(the middle part in Fig.  \ref{fig:r-bert}).
Wu and He \cite{wu-he-2019-enriching} exploit similar tokens to indicate entities' positions 
and predict relations by the concatenation of the hidden state of {[CLS]} 
and two averaged entity representations  (the rightmost part in Fig.  \ref{fig:r-bert}).
The method is named as {R-BERT}.

As illustrated in Fig. \ref{fig:r-bert}, EM-C can be regarded as a base model. 
EM-ES and R-BERT can be treated as the upgraded versions of EM-C.
Please refer to the original papers \cite{baldini-soares-etal-2019-matching,wu-he-2019-enriching} for more details.
The performance of EM-C, EM-ES, and R-BERT is evaluated and compared with each other under RDR in this paper. 
Note that these models (EM-C, EM-ES, and R-BERT) are not interpretable since they are very deep neural networks.

{
These models based on CNN, RNN, GCN, or pretrained language models are traditionally evaluated under relation classification. 
Under traditional evaluation of relation classification, the directionality of relations is considered by regarding a relation type with opposite directions as two class labels. 
However, the directionality of relations is lost since  
these two labels no longer reflect the directionality of relations. 
In this paper, the proposed evaluation, RDR, serves as a probe to 
inspect the capability of a model in recognizing  the directionality of relations.
With the help of RDR, one can further understand those black-box models that lack interpretability.

To clear up misunderstanding and  highlight the novelty and contributions of this paper, 
hypernymy classification \cite{nguyen-etal-2017-hierarchical,wang-etal-2019-improving} and relation classification   is distinguished  in terms of the directionality of relations. 
As for the directionality of a relation, the special relation \textit{IS-A} is studied in 
hypernymy classification \cite{nguyen-etal-2017-hierarchical,wang-etal-2019-improving}, a binary classification, 
which determines whether the only relation  \textit{IS-A} exists between an instance (or concept) and a concept.
The directionality of the relation \textit{IS-A} is considered to judge whether the instance (concept) is hyponym (hypernym).
In this paper, the directionality of arbitrary relations is further investigated in relation classification, which is a multi-class classification. 
}

\section{Relation Direction Recognition}

To distinguish the directionality of arbitrary relations, 
it is necessary for a model to both discern the relation between entities and capture the relative order of entities in a sentence.
From this perspective, it is a higher requirement for a model to recognize the directionality of relations.
If a model can recognize both the semantics of a sentence and  the order of its entities to predict a relation, 
the inverse relation should be predicted by the model on a similar sentence 
with the opposite entity annotation order. 
For example, a model should predict simultaneously the first sample as \textit{Father-Child({e1},{e2})} 
and the second one as \textit{Father-Child({e2},{e1})} in Table \ref{tab:example},
if it is able to recognize the relative order of entities.
Based on  this assumption, 
the difference or consistency of the two predictions can be compared to 
judge whether the model recognizes the directionality of a relation.

Before introducing RDR, some concepts need to be defined for convenience. 
Paired samples are two samples about relation classification that have the same sentence yet different annotations so that the two samples have different labels. Wherein one of them is called as the paired-sample of the other. 
For example, the two samples in Table \ref{tab:example} are called as a pair of paired-samples. 
Two sets are defined as paired sets 
if any sample in one of the sets can find its paired-sample in the other set and vice versa.
Similarly, one of them is called as the paired set of the other.
Formally, RDR is defined as follows.

\noindent \textbf{Definition 1.} 
\textit{Relation Direction Recognition (RDR) reflects  the difference or consistency  of two performance that a trained model achieves on a pair of 
	paired test 
	sets\footnote{Paired \textit{test}  sets indicate that the paired sets are used for \textit{test}. 
		Similarly, paired \textit{trainging}  sets used in the following parts indicate that the paired sets are used for \textit{trainging}. 
}}.


The above definition is a general description of RDR.
Next, three metrics are proposed to measure a model's performance under RDR 
from multiple angles.

\subsection{Measure Description}


\subsubsection{Performance Difference} 
The first metric comes from the direct comparison between a model's performance on a pair of paired test sets.
A model is supposed to perform similarly in a pair of paired test sets, 
if it can  recognize both the semantics of a sentence and  the order of its entities to predict a relation,
Otherwise, the model might show a big performance difference on the paired test sets.
Based on this assumption, the first metric, Performance Difference (PD), is proposed for RDR.
PD is defined as the difference of a model's performance between a pair of paired test sets (denoted by \textit{A} and \textit{B}, respectively), 
and is formalized as:
\begin{equation}
	PD	= |P_A-P_B|,
\end{equation}
where $P_A$ ($P_B$) is the performance (e.g. in the metric of Macro-F1) of a model in the test set \textit{A} (\textit{B}) and $|x|$ returns the absolute value of $x$.
Clearly, the larger PD, the worse a model recognizes the directionality of relations, 
since a model performs very differently on paired test sets.
Theoretically, the range of PD is from 0 to 1 (both boundary values included).

Obviously, PD originates from the comparison of a model in a traditional metric, 
which is usually an overall indicator. 
Therefore, 
through the difference of the overall performance of a model in paired test sets,
PD can implicitly reflect the degree to which the model recognizes the directionality of relations.

\subsubsection{Predictive Immobility Rate} 
If a model does not capture the relative order of entities 
and only depends on the semantics of a sentence to discern a relation, 
it might predict the same relation for a pair of paired-samples, 
since the paired-samples contain similar sentences.
From this perspective, 
the second metric, named Predictive Immobility Rate (PIR), is proposed to measure a model's performance for RDR.

%
Formally, PIR in a pair of paired test sets \textit{A} and \textit{B} is defined as follows:
\begin{equation}
	PIR = \frac{\sum_{i=1}^N{I(y_i^A = y_i^B \quad {\rm and} \quad y_i^X = t_i^X)}}{\sum_{i=1}^N{I(y_i^X = t_i^X)}},
\end{equation}
where \textit{N} is the number of samples in the test set {A} (B) and $X$ denotes the test set \textit{A} if a model performs better on \textit{A} than on \textit{B} and the test set \textit{B} otherwise. 
$y_i^A$ ($y_i^B$) and  $t_i^A$ ($t_i^B$) are the prediction relation and the ground-truth relation  for the $i$th 
sample\footnote{It is supposed  that the sample orders of a pair of paired test sets are similar.
	Namely, the \textit{i}th sample of one test set is the paired-sample of the \textit{i}th sample of the other test set.}
in the test set \textit{A} (\textit{B}), respectively. 
\textit{I (expression)} is an indicator function that returns 1 when \textit{expression} is true and 0 otherwise.


From the perspective of paired-samples, PIR directly reflects the consistency of predictions on pair-samples. 
Since pair-samples have inverse relations, the larger PIR, the worse a model recognizes the directionality of relations. 
Theoretically, the range of PIR is from 0 to 1 (both boundary values included). 

To recap, PD and PIR are both negative metrics that mean smaller is better. 
PD is an indirect indicator and dependent on the traditional metric, 
while PIR is a direct indicator and computed on the paired test sets. 
Analytically, PD only reflects the performance of a model in recognizing the directionality of relations. 
By contrast, PIR can further point out to what extent a model recognizes relations solely on the semantics of sentences. 
Therefore,  the understanding of a model can be deepened through PIR.

\subsubsection{Paired Predictive Rate} 
Conversely, if a model predicts simultaneously the correct relations
for a pair of paired-samples,
an inference can be insisted that 
the model can recognize the  directionality of relations.
In view of this hypothesis, 
the third metric is proposed to measure a model's ability to recognize the directionality of relations.
It is called as Paired Predictive Rate (PPR). 

Formally, PPR in a pair of paired test sets \textit{A} and \textit{B} is defined as follows:
\begin{equation}
	PPR = \frac{\sum_{i=1}^N{I(y_i^A = t_i^A \quad {\rm and} \quad y_i^B = t_i^B)}}{N}.
\end{equation}
Clearly, the larger PPR, the better a model recognizes the directionality of relations,
since PPR reflects a model's ability to predict both a relation and  its directionality. 
PPR is a positive indicator by definition.
Theoretically, the range of PPR is from 0 to 1 (both boundary values included).

Unlike PD, PIR and PPR are computed from the direct comparison between the predictions on a pair of paired test sets.
Therefore, the capability of a model in recognizing the directionality of relations could be explicitly reported by PIR and PPR. 



\section{Experiments}

\subsection{Dataset Preparation}

{SemEval-2010 Task 8}\footnote{\url{http://semeval2.fbk.eu/semeval2.php?location=data}}
(SemEval) \cite{hendrickx-etal-2010-semeval} 
often refers to a dataset for relation classification. 
SemEval contains 18 semantic relations and an artificial relation \textit{Other}.
The 18 semantic relations can be treated as 9 pairs of relations with inverse  directions.
\textit{Other} is defined as any relation beyond the aforementioned 18 relations. 
Officially, SemEval is split into a training set with 8,000 samples and a test set with 2,717 samples. 
Every sample in SemEval consists of a sentence with two annotated entities and 
a labeled relation that indicates the relationship between the entities.
Namely, the samples of SemEval are similar to the examples in Table \ref{tab:example}.
The official Macro-F1 score is reported on this dataset.

The paired set of the test set of SemEval is created for 
RDR\footnote{The dataset for RDR is available on \url{https://github.com/Saintfe/RDR}.}.  
For convenience, the original test set of SemEval is denoted by \textit{A} while the newly created test set is denoted by \textit{B}.
Namely, each example in the \textit{A} set can match its paired-sample in the \textit{B}  set and vice versa.
The paired sets (\textit{A} and \textit{B}) are used to evaluate a model for RDR.


Note that \textit{Other} is a special relation in SemEval.
If a sample is labeled by \textit{Other} in \textit{A}, its paired-sample is also labeled by \textit{Other} in B 
according to the definition of \textit{Other}. In other words, no relational directionality exists in \textit{Other}, 
so that it is hard to judge whether a model distinguishes the directionality of \textit{Other}.
Therefore, the samples labeled by \textit{Other} are not taken into account when PIR and PPR are reported.
Note that the official Macro-F1 also excludes \textit{Other}.

\subsection{Evaluated Model}
As mentioned in Related Work, 
EM-C is first evaluated as a baseline under RDR in this section,  
and EM-ES and R-BERT are evaluated for comparison in the next section.
EM-C only designates the positions of entities and 
utilizes the hidden state of [CLS] as  the final representation to predict a relation.
In other words, EM-C does not explicitly exploit the information of entities.

In addition, 
the method EM-C is based on BERT. 
There are two available pretrained model sizes of BERT: \textbf{BERT$_{ \bm{{\rm base}}}$} and 
\textbf{BERT$_{ \bm{ {\rm large}}}$}\footnote{ 
	In the following parts, 
	the subscripts \textit{base} and \textit{large} with  the italic format are used to indicate the two pretrained models \textbf{BERT$_{\rm base}$}, \textbf{BERT$_{\rm large}$}, respectively.
} 
from the HuggingFace Transformers 
Library\footnote{https://github.com/huggingface/transformers} 
\cite{wolf2019huggingface}.
Therefore, EM-C (\textit{base}) and EM-C (\textit{large}) are both 
trained and evaluated under RDR.

For both EM-C (\textit{base}) and EM-C (\textit{large}), we set the learning rate as 2e-5 and fine-tune them for 5 epochs. 
The batch size\footnote{
	The batch sizes are dependent on the model sizes.
	All models in this paper are  trained on two Nvidia GTX 1080Ti graphic cards. 
}
of EM-C (\textit{base}) is set as 100 while 
the batch size of  EM-C (\textit{large}) is set as 20.

\begin{table}[t]
	\centering	
	\caption{PD, PIR, and PPR of EM-C with two pretrained model sizes on the test set (\textit{A}) of SemEval and its paired test set (\textit{B}).}
	\begin{tabular}{llp{0.7cm}p{0.7cm}p{0.7cm}p{0.7cm}p{0.7cm}}
		\toprule
		Method & Size & \textit{A}  & \textit{B}  & PD  & PIR  & PPR  \\
		\midrule
		EM-C&base 	 	 & 81.27 	&7.66		& 73.61 & 79.03 &3.27\\		
		EM-C&large 	 	 & 89.00 	&15.96		& 73.04 & 65.77 &12.95\\
		\bottomrule
	\end{tabular}
	\label{tab:result-em-c}
\end{table}

\subsection{Results of EM-C under RDR}

PD, PIR, and PPR of EM-C (\textit{base}/\textit{large}) are reported on  the test set (i.e. \textit{A}) of SemEval and its paired test set (i.e. \textit{B}) in Table \ref{tab:result-em-c}.
For the method EM-C (\textit{base}), it obtains 81.27\% Macro-F1 on the test set \textit{A} while it only achieves 7.66\% Macro-F1 on \textit{B}. 
An absolute 73.61\% performance difference (PD) implicitly indicates EM-C (\textit{base}) may not recognize the directionality of relations.
Furthermore, EM-C (\textit{base}) achieves 79.03\% PIR that shows most paired-samples are predicted similarly on \textit{A} and \textit{B}. 
A high PIR indicates that EM-C  (\textit{base})  predicts relations mainly by the semantics of sentences without capturing the relative order of entities.
EM-C (\textit{base}) achieves 3.27\% PPR that indicates a few paired-samples can be predicted correctly at the same time.
The large PIR (79.03\%) and the small PPR (3.27\%) 
explicitly indicate that EM-C (\textit{base}) can hardly recognize the directionality of relations.
The reason for these poor results may be  that the design of EM-C does not explicitly utilize the information of entities.

When the larger pretrained model size is equipped, 
EM-C (\textit{large}) simultaneously achieves  better performance on \textit{A} (89.00\%) and \textit{B} (15.96\%) 
than  EM-C (\textit{base}).
Especially, EM-C (\textit{large}) obtains smaller PD (73.04\%), PIR (65.77\%), and larger PPR (12.95\%) than EM-C (\textit{base}), respectively.
These experimental results indicate that a larger pretrained model size can make EM-C better 
recognize the directionality of relations.

\section{Discussion}

To improve the ability of a model in recognizing the directionality of relations, 
some explorations are conducted from the perspective of model design or training.

\begin{table}[t]
	\centering	
	\caption{PD, PIR, and PPR of EM-ES and R-BERT with two pretrained model sizes on the test set (\textit{A}) of SemEval and its paired test set (\textit{B}).}
	\begin{tabular}{llp{0.7cm}p{0.7cm}p{0.7cm}p{0.7cm}p{0.7cm}}
		\toprule
		Method & Size & \textit{A}  & \textit{B}  & PD  & PIR  & PPR  \\
		\midrule
		EM-ES&\textit{base} 	 	 & 86.46 	&40.38		& 46.08 & 18.17 &33.94\\
		R-BERT&\textit{base} 	 & 86.81 	&46.38		& 40.43 & 8.66  &42.33\\		
		\midrule
		EM-ES&\textit{large} 	 & {89.23} 	&61.39		& 27.84 & {5.75}  &57.36\\
		R-BERT&\textit{large} 	 & 89.13 	&{62.11}		& 27.02 & {5.70} &59.21\\
		\bottomrule
	\end{tabular}
	\label{tab:result-em-es-r-bert}
\end{table}

\subsection{Model Design for Directionality of Relations}
There are many ways to make a method recognize the directionality of relations.
One of them is to let a method explicitly apply entities' information for relation classification. 
As mentioned in Related Work, 
EM-ES \cite{baldini-soares-etal-2019-matching} explicitly utilizes the entities' information by the concatenation of the hidden states of two\textit{ entity start makers}.
R-BERT \cite{wu-he-2019-enriching}  explores  both entities' information by using mean-pooling on the hidden states of entities and the sentence information by the hidden state of [CLS].

Similarly, EM-ES and R-BERT with two pretrained model sizes  are evaluated under RDR.
The hyper-parameters of  EM-ES and R-BERT with the \textit{base} (\textit{large}) pretrained model size  
are the same as those of EM-C with the \textit{base} (\textit{large}) pretrained model size, respectively.

Table \ref{tab:result-em-es-r-bert} reports the performance of  EM-ES (\textit{base}/\textit{large}) and R-BERT (\textit{base}/\textit{large}) for RDR.
EM-ES and R-BERT consistently outperform EM-C under  the \textit{base} and \textit{large} pretrained model sizes, respectively.
Take EM-ES (\textit{base})  as an example.
EM-ES (\textit{base}) achieves better performance on \textit{A} (86.46\%) and \textit{B} (40.38\%) than  EM-C (\textit{base}).
It indicates that entities' information is important for a sentence in relation classification.
One can observe that PD of EM-EC(\textit{base}) is 46.38\%,  much smaller than 73.61\% PD of EM-C (\textit{base}).
Meanwhile, 
EM-ES (\textit{base}) achieves 18.17\% PIR, significantly outperforming EM-C (\textit{base}). 
It achieves 33.94\% PIR, much better than EM-C (\textit{base}). 
Smaller PD, PIR, and larger PPR of  EM-ES (\textit{base}) than EM-C (\textit{base}) indicate that  
entities' information is also critical to recognize the directionality of relations.

Better performance is achieved on \textit{A} (86.81\%) and \textit{B} (46.38\%) by R-BERT (\textit{base})  than  EM-ES (\textit{base}).
It indicates that the sentence information can assist entities' information for relation classification 
when the pretrained model size is smaller.
Meanwhile, R-BERT (\textit{base}) achieves 40.43\% PD and 8.66\% PIR, which are smaller than 46.08\% PD and 18.17\% PIR of EM-ES (\textit{base}), respectively. 
R-BERT (\textit{base}) obtains 42.33\% PPR, larger than 33.94\% PPR of EM-ES (\textit{base}).
The evidence indicates that the sentence information is useful to recognize the directionality of relations 
when the size of pretrained model is smaller.

When the larger pretrained model size (\textit{large}) is equipped, 
EM-ES (\textit{large}) and R-BERT (\textit{large}) simultaneously achieve better performance than their \textit{base} versions, respectively.
Take EM-ES (\textit{large}) as an example. 
It achieves 89.23\% and 61.39\% on \textit{A} and \textit{B}, respectively.
The performance of EM-ES (\textit{large}) is better than  EM-ES (\textit{base}).
Especially, EM-ES (\textit{large}) obtains smaller PD (27.84\%), PIR (5.75\%), and larger PPR (57.36\%) than EM-ES (\textit{base}).
Similarly,  R-BERT (\textit{large}) performs better than R-BERT (\textit{base}) 
in PD, PIR, and PPR, respectively.
These results  indicate  that a larger pretrained model size can make these methods better 
recognize the directionality of relations.
Besides, EM-ES (\textit{large}) and R-BERT (\textit{large}) are comparative in the test set \textit{A}.

\subsection{Encode Directionality of Relations into a Model}

\begin{figure}[t]
	\centering
	\includegraphics[width=\columnwidth]{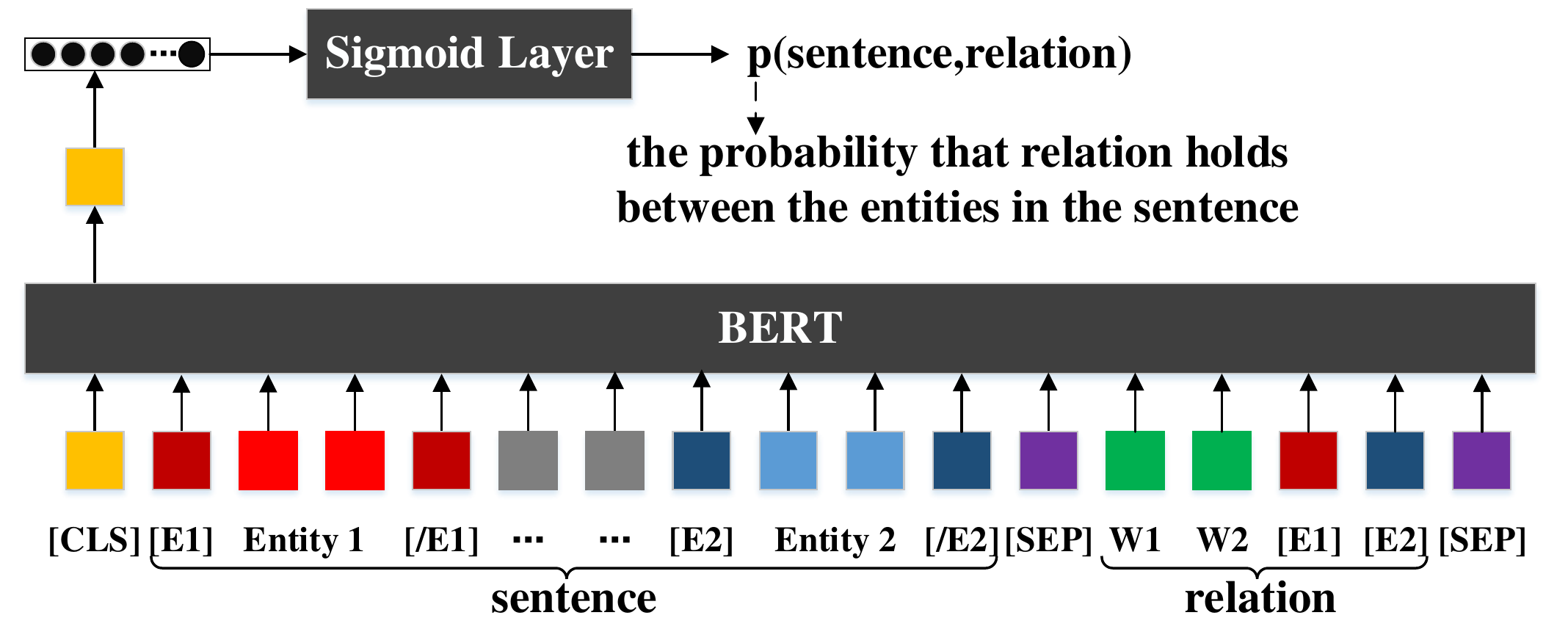}
	\caption{The architecture of the proposed method \textit{EMR-C}. 
		W1 and W2 denote Cause and Effect respectively when the relation is Cause-Effect(e1,e2).}
	\label{fig:emr-c}
\end{figure}

A relation is usually literally expressed as a text sequence with two words and two entity symbols, for example, \textit{Cause-Effect(e1,e2)}.
The relation can be literally processed  as a text sequence ``Cause Effect e1 e2'', which is both semantic and directional. 
Intuitively,  the literal texts of relations are encoded into a model, which is useful for the model to recognize the directionality of relations.

Based on this intuition, a variant of EM-C, called EMR-C, is proposed to encode the information of literal relations to learn the directionality of relations.
Specifically,
EMR-C takes a literal relation and a sentence with entity markers as inputs, as illustrated in Fig.  \ref{fig:emr-c}.
EMR-C outputs the hidden state of [CLS] into a binary classifier (i.e. the sigmoid layer) to obtain a probability 
that the relation holds between the entities in the sentence.

In addition to the difference in inputs, EMR-C learns a binary classifier while EM-C does a multi-class classifier.
Since SemEval is designed for multi-class classifiers, the dataset can not be directly used to train and evaluate EMR-C.
Therefore, SemEval needs to be adjusted to fit the  binary scenario.

In the training phase, when a sentence with two entities and its relation are combined to be fed into EMR-C as inputs, 
the combination of the sentence and its labeled relation forms a positive sample for EMR-C,  
since the relation holds between the entities of the sentence. 
In this way, every sample in the training set of SemEval can form one positive sample for EMR-C.
As to negative samples for EMR-C, the sentence of every  sample in the training set remains unchanged and is combined with  
a wrong relation (randomly selected) to serve as a negative sample.

In the test phase, for a sample of the test set,  
the combination of its sentence and each of the candidate relations is firstly fed into EMR-C. 
Secondly, EMR-C obtains the probability of the combination between the sentence and each of the candidate relations. 
Finally,  EMR-C chooses the relation with the maximum probability as the predicted relation for the sample.
Formally, the predicted relation (denoted by $y$) is chosen for a sentence (denoted by $s$) with two entities by 
\begin{equation}
	\label{eq:emr-c}
	y =  \argmax_{r \in R} p(s, r),
\end{equation}
where $R$ is the set of the candidate relations and $p(s,r)$ denotes the probability 
that  the \textbf{\underline{r}}elation ($r$) holds between the entities in the \textbf{\underline{s}}entence ($s$).


Similarly, EMR-C with two pretrained model sizes  are evaluated under RDR.
The hyper-parameters of EMR-C (\textit{base}) and  EMR-C (\textit{large})  
are the same as those of EM-C (\textit{base}) and  EM-C (\textit{large}), respectively.

\begin{table}[t]
	\centering	
	\caption{PD, PIR, and PPR of EMR-C with two pretrained model sizes  on the test set (\textit{A}) of SemEval and its paired test set (\textit{B}).}
	\begin{tabular}{llp{0.7cm}p{0.7cm}p{0.7cm}p{0.7cm}p{0.7cm}}
		\toprule
		Method & Size & \textit{A}  & \textit{B}  & PD  & PIR  & PPR  \\
		\midrule
		EMR-C&\textit{base} 	 	 & 77.32 	&23.59		& 53.73 & 35.31 &22.45\\		
		EMR-C&\textit{large} 	 	 & 84.02 	&46.10		& 37.92 & 8.39  &39.20\\		
		\bottomrule
	\end{tabular}
	\label{tab:result-emr-c}
\end{table}

Table \ref{tab:result-emr-c} presents PD, PIR, and PPR of EMR-C (\textit{base}/\textit{large}).
EMR-C (\textit{base}/\textit{large}) achieve worse performance than EM-C (\textit{base}/\textit{large})  on \textit{A}, respectively, 
which may result from the mismatch between the  binary design in EMR-C and the multi-class situation in SemEval. 
On the contrary, EMR-C (\textit{base}/\textit{large}) achieve better performance than EM-C \textit{base}/\textit{large})  on \textit{B}, respectively. 
Meanwhile, smaller PD,  PPR,  and larger PPR are obtained by EMR-C (\textit{base}/\textit{large}) than EM-C (\textit{base}/\textit{large}). 
These experimental results show that EMR-C performs better than EM-C in recognizing the directionality of relations.
Therefore, the evidence further indicates that  the literal texts of relations are useful to recognize the directionality of relations.


Besides, this solution is not recommended for two reasons.
On one hand, it is not logically straightforward and too cumbersome. The original problem is a multi-class classification. Then it is transformed into an equivalent binary classification and is solved by a binary method. Finally, the method is evaluated by a multi-class metric by the specific post-processing procedure in Eq. \ref{eq:emr-c}. The whole solution is a little indirect. On the other hand, 
it requires more time and computing resources. When the original problem is transformed into a binary classification, there are no ready-made negative samples available. 
Therefore, negative samples with the same number of positive samples need to be sampled.
This results in twice the number of original samples to be trained with the consumption of more time and computing resources.
On the whole, EMR-C is just a compromise with shortcomings for recognizing the directionality of relations.

\subsection{Model Training for Directionality of Relations}

\begin{figure*}[htbp]
	\centering
	\subfigure[EM-C (\textit{base})]{
		\begin{minipage}[t]{0.33\linewidth}
			\centering
			\includegraphics[width=\linewidth]{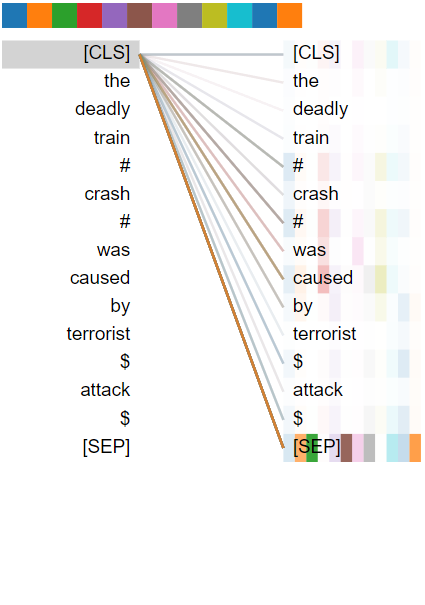}\\
			\vspace{0.02cm}
		\end{minipage}%
	}%
	\subfigure[EM-C-M (\textit{base})]{
		\begin{minipage}[t]{0.33\linewidth}
			\centering
			\includegraphics[width=\linewidth]{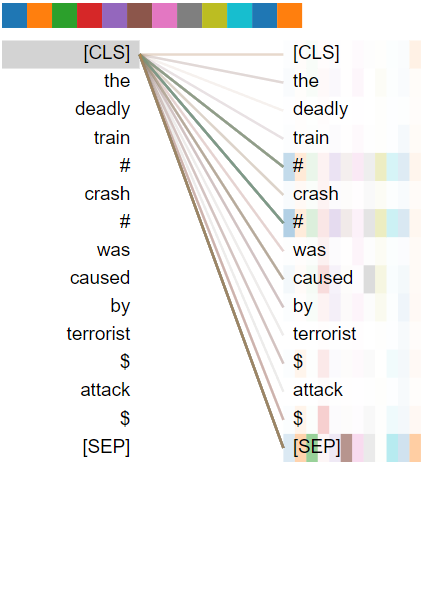}\\
			\vspace{0.02cm}
		\end{minipage}%
	}%
	\subfigure[EMR-C (\textit{base})]{
		\begin{minipage}[t]{0.33\linewidth}
			\centering
			\includegraphics[width=\linewidth]{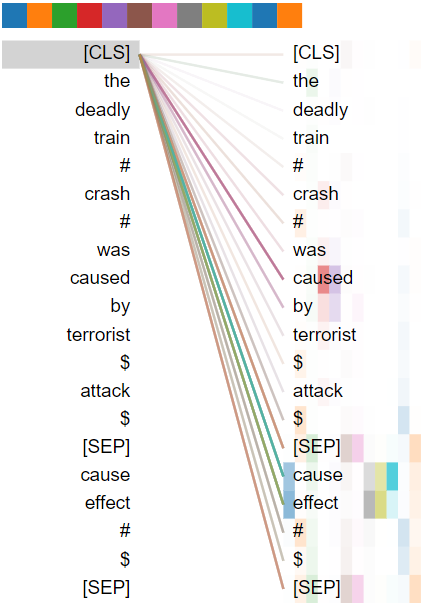}\\
			\vspace{0.02cm}
		\end{minipage}%
	}%
	\centering
	\caption{
		Multi-head attention of EM-C (\textit{base}, the 9th layer), EM-C-M (\textit{base}, the 9th layer), 
		and EMR-C (\textit{base}, the 5th layer)\textsuperscript{9}
		that [CLS] attends to words in the sentence ``\textit{the deadly train \textless e1\textgreater \, crash \textless/e1\textgreater \, was caused by terrorist \textless e2\textgreater \,  attack \textless/e2\textgreater}''.
		The sentence is labeled as \textit{Cause-Effect(e1,e2) } in the test set of SemEval.
		Entity markers \textless e1\textgreater \, and \textless/e1\textgreater \, are placed by \# while \textless e2\textgreater \, and \textless/e2\textgreater \, are replaced by \$.
		Attention weights in multiple heads are presented in various colors. Color opacity indicates the value of an attention weight.
	}
	\vspace{-0.2cm}
	\label{fig:attention}
\end{figure*}

Alternatively,  in the training phase, a model can be forced to learn the directionality of relations 
by constructing paired-samples in the training set.
Specifically, the paired training set is created from the original training set of SemEval. 
Then, the original training set and its paired set are combined into a merged training set. 
As a result, paired-samples appear in the merged training set. 
Trained by the merged training set, a model has to learn the directionality of relations, 
since a pair of paired-samples contain similar sentences yet reverse relations (except for \textit{Other}).

Note that the merged training set  can be treated as a data augmentation technique that is similar to \cite{xu-etal-2016-improved}. 
However, the motivation behind forming the merging set is quite different. 
Xu et al. \cite{xu-etal-2016-improved} augment training data in order to train a deeper RNN, which avoids overfitting, 
while we construct paired-samples in the training set to force a model to learn the directionality of relations.

EM-C is first trained with the \textit{base} pretrained model size on the merge training set. 
For convenience, the model trained on the \textbf{\underline{m}}erged training set is denoted by EM-C-M.
Similarly, the hyper-parameters of  EM-C-M (\textit{base}) are the same as those of EM-C (\textit{base}).

\begin{table}[t]
	\centering
	\caption{PD, PIR, and PPR of EM-C-M with the pretrained model size \textit{base}  on the test set (\textit{A}) of SemEval and its paired test set (\textit{B}).}
	\begin{tabular}{llp{0.7cm}p{0.7cm}p{0.7cm}p{0.6cm}p{0.6cm}}
		\toprule
		Method & Size & \textit{A}  & \textit{B} & PD & PIR & PPR 	\\
		\midrule
		EM-C-M&\textit{base} &87.85 &87.58 &0.27 & 0.25& 87.72 \\
		\bottomrule
	\end{tabular}
	\label{tab:result-em-c-m}
\end{table}

The results of EM-C-M (\textit{base}) are shown in Table \ref{tab:result-em-c-m}.
EM-C-M (\textit{base}) achieves a significant  increase of absolute 6.58\%  over EM-C (\textit{base}) on the test set \textit{A} (87.85\% vs 81.27\%).
Especially, the improvement of EM-C-M (\textit{base}) on \textit{B}  is very significant over  EM-C (\textit{base}), reaching an absolute 79.92\% increase  (87.58\% vs 7.66\%).
More importantly,  EM-C-M (\textit{base}) obtains a huge decrease of PD with  absolute 78.76\% over EM-C (\textit{base}) (0.27\% vs 79.03\%).
PIR of EM-C-M (\textit{base}) drops to 0.25\% from 79.03\% PIR of EM-C (\textit{base}).
Meanwhile, EM-C-M (\textit{base}) achieves 87.72\% PPR that is much larger than 3.27\% of EM-C (\textit{base}).
The experimental results significantly indicate 
that it is effective for a model to recognize the directionality of relations by constructing paired-samples in the training set.

\subsection{Case Study}
\label{subsec:case-study}

It is hard to interpret why or not  these models recognize the directionality of relations.
There is a way to understand the decision of a model by exploiting attention that can obtain a distribution over 
inputs \cite{ghaeini-etal-2018-interpreting}.
Notably, the effectiveness of attention is controversial for interpretability.
Some researchers argue that the attention mechanism  can help the understanding of deep models \cite{vaswani2017attention}.
The others claim that attention of a deep model is by no means a fail-safe indicator to  represent input components’ overall importance \cite{serrano-smith-2019-attention}.
We argue that attention is still a choice to interpret deep models' decisions under the current research.
In this paper,  attention is exploited to further understand a model in recognizing  the directionality of relations.
A visualization tool\footnote{https://github.com/jessevig/bertviz/} 
\cite{vig-2019-multiscale} is utilized to illustrate how much weight a model allocates to various input elements (i.e. tokens).

Fig. \ref{fig:attention} visualizes multi-head attention that [CLS] attends to the words of a sample from  EM-C (\textit{base}), EM-C-M (\textit{base}), and EMR-C (\textit{base}).
These words are differently colored according to their obtained attention weights from [CLS] whose hidden state is used for classification in the three methods.
From this figure, we can obverse that 1) EM-C (\textit{base}) mostly focuses on the key word ``\textit{caused}'' to recognize the relation \textit{Cause-Effect(e1,e2).}
However, EM-C (\textit{base}) is hard to recognize the directionality of the relation even though it captures the most important word in the sentence.
2) EM-C-M (\textit{base}) concentrates both on the entity marker \textit{\#} and the key word ``\textit{caused}" 
so that it can capture the meaning of the relation and  recognize its directionality.
3) EMR-C (\textit{base}) attaches more importance to the key word ``\textit{caused}" and the literal relation ``\textit{cause effect}" 
to make it discern the meaning of the relation.
Meanwhile, EMR-C (\textit{base}) allocates a small part of attention to the entity markers (i.e. \textit{\#} and \textit{\$}).
Therefore, it has a weak ability to recognize the directionality of relations.

\subsection{Relation Classification}

\begin{table}[t]
	\centering
	\caption{Performance (Macro-F1) comparison on the SemEval. 
		$\dagger$ marks results reported in the original papers.}
	\begin{tabular}{p{3.05cm}lp{0.5cm}p{0.5cm}p{0.5cm}p{0.5cm}p{0.7cm}}
		\toprule
		Method &  \textit{A}  & \textit{B} & PD & PIR & PPR 	\\
		\midrule
		SVM $\dagger$  & 82.2 & - & - & - & - \\
		\midrule
		PA-LSTM $\dagger$ & 82.7 & - & - & - & - \\
		SDP-LSTM $\dagger$ & 83.7 & - & - & - & - \\
		SPTree $\dagger$  & 84.4 & - & - & - & - \\
		C-GCN  & 84.79 & 25.15 & 59.64 & 3.20 & 24.17 \\
		C-AGGCN $\dagger$ & 85.7 & - & - & - & - \\
		Att-Pooling-CNN $\dagger$  & 88.0 & - & - & - & - \\
		\midrule
		Entity-Aware BERT $\dagger$ & 89.0 & - & - & - & - \\
		KnowBert-W+W $\dagger$ & 89.1 & - & - & - & - \\
		EM-ES $\dagger$ & 89.2 & - & - & - & - \\
		R-BERT $\dagger$ & 89.25 & - & - & - & - \\
		EM-ES-MTB $\dagger$ & 89.5 & - & - & - & - \\
		\midrule
		EM-C-M (\textit{large})  	 & 89.50		&89.53 & 0.03 &	0.05 & 90.01	\\
		R-BERT-M (\textit{large})  	 & 89.87 		&89.92 &0.05 &0 & 90.59\\
		EM-ES-M (\textit{large})  	 & \textbf{90.48}	    &90.21 & 0.27 & 0 & 91.43		\\
		\bottomrule
	\end{tabular}
	\label{tab:result-all}
\end{table}

Table \ref{tab:result-em-c-m} indicates that 
not only the directionality of relations can be learned by training a model on the merged training set with paired-samples,
but also  a significant improvement is achieved by exploiting  paired-samples.
Based on this observation, EM-C, EM-ES, and R-BERT are trained with the \textit{large} pretrained model size on the merged  training set, 
and the resulting models are denoted by EM-C-M, EM-ES-M, and R-BERT-M, respectively.
The hyper-parameters of EM-C-M (\textit{large}), EM-ES-M (\textit{large}), and R-BERT-M (\textit{large}) are the same as those of EM-C (\textit{large}).
The performance of these methods on the original test set \textit{A} is presented with the proposed metrics PD, PIR, PPR in Table \ref{tab:result-all}.

Many methods are compared with EM-C-M (\textit{large}), EM-ES-M (\textit{large}), and R-BERT-M (\textit{large}).
Baseline methods include: 1) SVM \cite{rink-harabagiu-2010-utd}, which classifies on traditional features,
2) PA-LSTM \cite{zhang-etal-2017-position}, which incorporates positions of entities into LSTM, 
3) SDP-LSTM \cite{xu-etal-2015-classifying}, which exploits the shortest dependency path (SDP) between entities by LSTM,
4) SP-Tree \cite{miwa-bansal-2016-end}, which applies LSTM and Tree-LSTM \cite{tai-etal-2015-improved} to SDP.
5) C-GCN \cite{zhang-etal-2018-graph}, which explores GCN to dependency trees of sentences, 
6) C-AGGCN \cite{zhang-etal-2018-graph}, which further introduces the attention mechanism into C-GCN, 
and 7) Att-Pooling-CNN \cite{wang-etal-2016-relation}, which proposes multi-level attention CNNs.
State-of-the-art methods include: 
1) Entity-Aware BERT \cite{wang-etal-2019-extracting}, which incorporates the entity-aware attention into BERT’s self-attention,
2) KnowBert-W+W \cite{peters-etal-2019-knowledge}, which integrates knowledge from both WordNet \cite{miller-1995-wordnet} and Wikipedia into BERT,
3) EM-ES,  4) R-BERT, 5) EM-ES-MTB\cite{baldini-soares-etal-2019-matching}, which adopts a learning method, named \textit{matching the blanks} (MTB), to train EM-ES.

\footnotetext[9]{We observe that top layers capture semantic features as \cite{jawahar-etal-2019-bert} when a single text is fed in BERT 
	while middle layers capture semantic features  when two texts are fed in BERT.}

From Table \ref{tab:result-all}, we can observe that 1) 
EM-ES-M (\textit{large}) achieves a new state-of-the-art on SemEval with 90.48\% 
Macro-F1\footnote[10]{Only published papers are considered and the preprint papers \cite{cohen-etal-2020-relation, li-tian-2020-downstream} have not yet passed peer review.}. 
Meanwhile, R-BERT-M (\textit{large}) and EM-C-M (\textit{large}) achieve competitive performance 
in the leaderboard\footnote[11]{\url{https://paperswithcode.com/sota/relation-extraction-on-semeval-2010-task-8}}
of SemEval. 
2) EM-C-M (\textit{large}) EM-ES-M (\textit{large}), and R-BERT-M (\textit{large}) consistently achieve better performance 
than EM-C (\textit{large}), EM-ES (\textit{large}), and R-BERT (\textit{large}),
whose results are reported in Table  \ref{tab:result-em-c} and Table \ref{tab:result-em-es-r-bert}, respectively.
These consistent results  indicate that paired-samples can train models more effectively.
3) Each of the three methods performs similarly on \textit{A} and \textit{B}, achieves very low PD and PIR, and obtains very high PPR.
These empirical results confirm  again that exploiting paired-samples to train models can not only force models to recognize the directionality of relations 
but also further improve the performance of models. 4) As a representative of baseline methods, C-GCN is evaluated under RDR. 
The large PD (59.64 \%) and the small PPR (24.17\%) indicate that C-GCN is weak in recognizing the directionality of relations.

Table \ref{tab:result-all} also indicates that PD and PIR are very low  under the training of paired-samples. 
The two negative  metrics are easy to close the boundary extreme (0) when models are capable of recognizing the directionality of relations, 
so that they are unsuitable for this situation. 
However, PPR still works well in this situation. 
Therefore, PPR is robust and recommended to be the metric for models that are able to recognize the directionality of relations. 
When models are weak in recognition of the relation directionality, PD and PIR can also be adopted.

\section{Conclusion}
In this paper, we propose a new evaluation task, Relation Direction Recognition (RDR), to 
explore the degree to which a model recognizes the directionality of relations.
RDR reflects the difference or consistency  of two performance that a trained model achieved on a pair of paired test sets.
Three metrics, which are Performance Difference (PD),  Predictive Immobility Rate (PIR), and Paired Predictive Rate (PPR), 
are proposed to measure the performance of a model for RDR.
Several state-of-the-art methods are evaluated under RDR.
Experimental results indicate that there are clear gaps among these methods 
even though they achieve similar performance in the traditional metric (i.e. Macro-F1).
A couple of suggestions from the perspectives of model design or training are discussed to make a model learn the directionality of relations.
From the perspective of model design, 
explicit usage of entities' information or literal texts of relations can improve the capability of a model in recognizing the directionality of relations.  
From the perspective of model training, 
exploiting paired-samples to train models can 
not only force models to recognize the directionality of relations 
but also further improve the performance of models. 

\bibliographystyle{IEEEtran}
\bibliography{sample-base}

\end{document}